An Initial Seed Selection Algorithm for K-means Clustering of Georeferenced Data to Improve Replicability of Cluster Assignments for Mapping Application

Fouad Khan

Central European University-Environmental Sciences and Policy Department

Nador utca 9, 1051 Budapest, Hungary

fouadmkhan@yahoo.com

Phone: +1 434 987 4922

Abstract

K-means is one of the most widely used clustering algorithms in various disciplines, especially for large datasets. However the method is known to be highly sensitive to initial seed selection of cluster centers. K-means++ has been proposed to overcome this problem and has been shown to have better accuracy and computational efficiency than k-means. In many clustering problems though –such as when classifying georeferenced data for mapping applications- standardization of clustering methodology, specifically, the ability to arrive at the same cluster assignment for every run of the method i.e. replicability of the methodology, may be of greater significance than any perceived measure of accuracy, especially when the solution is known to be non-unique, as in the case of k-means clustering. Here we propose a simple initial seed selection algorithm for k-means clustering along one attribute that draws initial cluster boundaries along the "deepest valleys" or greatest gaps in dataset. Thus, it incorporates a measure to maximize distance between consecutive cluster centers which augments the conventional k-means optimization for minimum distance between cluster center and cluster members. Unlike existing initialization methods, no additional parameters or degrees of freedom are introduced to the clustering algorithm. This improves the replicability of cluster assignments by as much as 100% over k-means and k-means++, virtually reducing the variance over different runs to zero, without introducing any additional parameters to the clustering process. Further, the proposed method is more computationally efficient than k-means++ and in some cases, more accurate.

1. Introduction

Clustering or classification of data into groups that represent some measure of homogeneity across a given variable range or values of multiple variables, is a much analyzed and studied problem in pattern recognition. K-means clustering is one of the most widely used methods for implementing a solution to this problem and for assigning data into clusters. The method in its initial formulation was first proposed by Mac Queen in 1967 (Mac Queen 1967) though the approximation developed by Lloyd in 1982 (Lloyd 1982) has proven to be most popular in application. The method assumes apriori knowledge of the number of clusters *k* and requires seeding with initial values of centers of these clusters in order to be implemented. These initial seed values have been shown to be an important determinant of the eventual assignment of data to clusters. In other words, k-means clustering is highly sensitive to the initial seed selection for the value of cluster centers (Peña, Lozano et al. 1999).

K-means++ has been proposed to overcome this problem and has been shown to produce a scale improvement in algorithm accuracy and computational efficiency or speed (Ostrovsky, Rabani et al. 2006; Arthur and Vassilvitskii 2007). The algorithm assesses the performance of the initial seed selection based on the sum of square difference between members of a cluster and the cluster center, normalized to data size. While this is a worthwhile means of assessing method performance, it may be noted that in many clustering applications, the replicability of the resultant cluster assignment can be much more desirable than the homogeneity of the cluster perceived through an objective measure.

We encountered one such application of the clustering problem while trying to cluster georeferenced data into classes for mapping and visualization through a Geographical Information Systems (GIS) software suite. Commercially available GIS software ArcGIS for instance utilizes a proprietary modification of Jenks natural breaks algorithm (Jenks 1967) to classify values of a variable for visualization in maps (ArcGIS 2009). The classification this method obtains seems to reproduce itself with remarkable accuracy for each run. The clustering bounds do not vary from run to run, even with variable values in eleven significant figures.

Jenks' algorithm differs only slightly from k-means clustering. K-means using Lloyd's algorithm aims to minimize the following cost function $C$ defined in Equation 2;

$$C = \sum_{\substack{1 \leq i \leq n \\ 1 < j < k}} dist(d_i, c_j) \qquad \text{Equation 2}$$

Where $n$ is the data size of number of data points, $k$ is the number of clusters and $dist(d_i, c_j)$ computes the Euclidean distance between point $d_i$ and its closest center $c_j$. The algorithm runs as follows;

a) Select centers $c_1,…,c_k$ at random from the data.
b) Calculate the minimum cost function $C$, assigning data points $d_1,…,d_n$ to their respective clusters having the closest mean.
c) Calculate new centers $c_1,…,c_k$ as means of the clusters assigned in step 2.
d) Repeats steps $b$ and $c$ until no change is observed in center values $c_1,…,c_k$.

Jenks algorithm differs in that instead of $C$ it minimizes the cost function $J$, defined in Equation 3;

$$J = C - \sum_{1 \leq j \leq (k-1)} dist(c_{j+1}, c_j) \qquad \text{Equation 3}$$

As can be seen in Equation 3, Jenks' algorithm not only searches for minimum distance between data points and centers of clusters they belong to but for maximum difference between cluster centers themselves (Jenks 1967).

If we are trying to develop a methodology for geo-processing; say a utility that studies the scaling characteristic of a city and models the distribution of sizes of housing within different size clusters, it can be essential to have a clustering mechanism that produces almost exactly similar results each time. Drawing inspiration from Jenks' algorithm, we propose an initial seed selection algorithm for k-means clustering that produces similar clusters on each run. We compare our results to those obtained by k-means as well as the widely used k-means++ initial seed selection methodology. K-means++ selects the initial centers as follows;

a) Select one center at random from dataset.
b) Calculate squared distance of each point from the nearest of all selected centers and sum the squared distances.
c) Choose the next center at random. Calculate sum of squared distances. Re-select this center and calculate the sum of squared distances again. Repeat a given 'number of trials' and select the center with the minimum sum of squared distance as the next center.
d) Repeat steps *b* and *c* until *k* centers are selected.

The methodology is novel in that unlike other initial seed selection algorithms, it does not introduce any new parameters (such as number of trials for k-means++) in the clustering algorithm thereby avoiding additional degrees of freedom. By clustering along the deepest valleys or highest gaps in the data series, the method introduces a measure of distance between cluster centers augmenting the k-means optimization for minimum distance between cluster center and cluster members. Additionally, unlike initialization algorithms like k-means++ there is no randomness involved in the algorithm and the initial clusters obtained are always the same.

2. Materials and Method

We propose the following method for calculating initial seed centers of k-means clustering along one attribute.

a) Sort the data points in terms of increasing magnitude $d_1,...,d_n$ such that $d_1$ has the minimum and $d_n$ has the maximum magnitude.

b) Calculate the Euclidean distances $D_i$ between consecutive points $d_i$ and $d_{i+1}$ as shown in Equation 4;

$$D_i = d_{i+1} - d_i; \quad \text{where } i = 1,..., (n\text{-}1) \qquad \text{Equation 4}$$

c) Sort $D$ in descending order without changing the index $i$ of each $D_i$. Identify $k$-1 index $i$ values $(i_1,...,i_{(k-1)})$ that correspond to the $k$-1 highest $Di$ values.

d) Sort $i_1,...i_{(k-1)}$ in ascending order. The set $(i_1,...,i_{(k-1)},i_k)$ now forms the set of indices of data values $d_i$, which serve as the upper bounds of clusters $1,...,k$; where; $i_k = n$.

e) The corresponding set of indices of data values $d_i$ which serve as the lower bounds of clusters $1,...,k$ would simply be defined as $(i_0, i_1+1, ..., i_{(k-1)}+1)$, where $i_0 = 1$.

f) The values of cluster centers $c$ will now simply calculated as the mean of $d_i$ values falling within the upper and lower bounds calculated above. This set of cluster centers $(c_1,...,c_k)$ will form the initial seed centers.

The methodology discussed above simply draws the cluster boundaries at points in the data where the gap between consecutive data values is the highest or the data has deepest 'valleys'. In this way, a measure of distance is brought between consecutive cluster centers.

The method can be easily implemented for small to medium size datasets by using the spreadsheet freely available for download at *http://ge.tt/api/1/files/7FON8KH/0/blob?download*.

To test the replicability of cluster assignments produced using this methodology, the same data was clustered using this methodology ten times. The variance observed in cluster centers for these ten runs was calculated and averaged over the number of cluster centers. For comparison similar analysis was performed employing k-means and widely used k-means++ initial seeding methodology and the variance averaged over number of cluster centers was calculated.

The analysis was run for five different datasets. The first data is the popular Iris dataset from UC Irvine Machine Learning Repository (UCIMLR) (Fisher 1936). Attribute one of the data was used for clustering. The data having 150 points was classed into 5 clusters. The second data is US census block wise population data for the Metropolitan Statistical

Area (MSA) of St. George Utah. The population, land area and water body area data was downloaded from the US Census Bureau website (US Census Bureau 2010). The area was calculated by summing and water and land areas for the census block. The population density for each census block was estimated by dividing population for the block with the area for the block. The data having 1450 points was clustered along population density into 10 clusters. The third data was Abalone dataset from UCIMLR (Nash, Sellers et al. 1994). Attribute 5 was used for clustering. The data has 4177 instances and was clustered into 25 classes. The fourth set of data was cloud cover data downloaded from Phillipe Collard (Collard 1989). Data in column 3 was used for cluster analysis. The data having 1024 points was clustered in 50 clusters. The fifth data set was randomly generated normally distributed data with mean 10 and standard deviation of 1. The data having 10,000 points was clustered into 100 clusters.

3. Results

While the objective of development of this method is to produce more replicable results, the sums of squared differences between cluster members and cluster centers between the proposed method and k-means++ were compared and are juxtaposed in Table 1. As can be seen in Table 1, k-means++ in general continues to produce more accurate clustering using this methodology, though for two of the five datasets, our method produced better results.

Table 1: Sum of Squared Differences between Cluster Members and their Closest Centers (Normalized to Data size)

| Dataset | k-means++ | Proposed method | Reduction% |
|---|---|---|---|
| **Iris** | 0.042243916 | 0.037471719 | 11.30% |
| **St. George** | 2.39419E-07 | 1.76868E-07 | 26.13% |
| **Abalone** | 0.000817549 | 0.001229598 | -50.40% |
| **Cloud** | 2.379979794 | 5.22916047 | -119.71% |
| **Normal** | 0.000644885 | 0.001465068 | -127.18% |

As shown in Table 2, our proposed method is also significantly faster than k-means++, clustering as much as 89% faster than k-means++ in some cases. The advantage in clustering speed is obtained over the initial seed selection, where k-means++ takes significantly longer comparative to both, our proposed method and k-means [2].

Table 2: Algorithm Running Time (Seconds)

| Dataset | k-means++ | Proposed method | Reduction% |
|---|---|---|---|
| **Iris** | 0.101 | 0.011 | 89.11% |
| **St. George** | 2.312999994 | 0.438000001 | 81.06% |
| **Abalone** | 19.79400002 | 16.191 | 18.20% |
| **Cloud** | 7.771000001 | 1.886000005 | 75.73% |
| **Normal** | 207.8150008 | 145.3870012 | 30.04% |

The premier advantage of our proposed method over k-means and k-means++ though is in improving method replicability. The results are presented in Table 3. As can be seen in all three cases, the variance was virtually reduced to zero using our method, which was at least a 90% improvement on k-means++ and k-means.

Table 3: Variance of Centers over Ten (10) Runs Averaged to Number of Clusters

| Dataset | Proposed method | k-means++ | Reduction% | k-means | Reduction% |
|---|---|---|---|---|---|
| **Iris** | 4.73317E-31 | 0.046361574 | 100.00% | 0.499704 | 100.00% |
| **St. George** | 1.12847E-37 | 1.22722E-36 | 90.80% | 1.23E-36 | 90.80% |
| **Abalone** | 2.37968E-32 | 0.003285155 | 100.00% | 0.005395 | 100.00% |
| **Cloud** | 1.72981E-28 | 31.54401321 | 100.00% | 22.24461 | 100.00% |
| **Normal** | 5.75868E-31 | 0.009478013 | 100.00% | 0.054631 | 100.00% |

4. Discussions and Conclusion

The method for initial seed selection of algorithm we propose reduces the variance of clustering to zero accurate up to eleven significant figures, for clustering along one attribute or dimension. The further advantage of the proposed initialization method is that unlike k-means++ it does not introduce any new variables within the analysis, such as the number of trials. Almost perfect replicability and avoidance of additional degrees of freedom make the method especially suited for inclusion as part in a protocol or standard methodology or algorithm. Further, the method also produces results faster than k-means++ and hence is more computationally efficient at least in two-dimensional space. The method has applications in all areas of data analysis where a Jenks style 'natural' classification, with a high level of replicability may be needed. It has the following distinct advantages over other initialization methods and naked k-means implementation;

- The results are highly replicable
- The method is fast and easy to implement
- No additional degrees of freedom or modifiable parameters are introduced that may need expert input for getting replicable results

- The clustering may be more 'natural' in the manner of Jenks' algorithm considering that a measure of distance between cluster centers is introduced to augment the k-means optimization of minimum distance between cluster members and cluster center.

Above advantages can render the initialization method highly useful in all areas where large datasets have to be handled or a 'natural' classification of data is sought. This includes areas like bioanalysis for instance where density based clustering is commonly deployed; the method can be made part of a more detailed analysis regime with confidence that the replicability of the results will not be negatively affected by the clustering algorithm. In the area of market segmentation and computer vision, the method can be used to standardize clustering results. This makes the method especially suited to utility development for GIS applications.